\documentclass{article} % For LaTeX2e
\usepackage{iclr2026_conference,times}

\usepackage{amsmath,amsfonts,bm}

\def\eqref#1{equation~\ref{#1}}
\def\1{\bm{1}}

\DeclareMathAlphabet{\mathsfit}{\encodingdefault}{\sfdefault}{m}{sl}
\SetMathAlphabet{\mathsfit}{bold}{\encodingdefault}{\sfdefault}{bx}{n}

\usepackage{hyperref}
\usepackage{cleveref}
\usepackage{url}
\usepackage{xcolor}
\usepackage{booktabs}
\usepackage{multirow}
\usepackage{multicol}
\usepackage{tablefootnote}
\usepackage[pdftex]{graphicx}

\title{SIGMark: Scalable In-Generation Watermark with Blind Extraction for Video Diffusion}

\iclrfinalcopy

\makeatletter
\def\@maketitle{\vbox{\hsize\textwidth
{\LARGE\sc \@title\par}
\ificlrfinal
  \lhead{Published as a conference paper at ICLR 2026}
  \def\And{\end{tabular}\hfil\linebreak[0]\hfil
          \begin{tabular}[t]{c}\rule{\z@}{24pt}\ignorespaces}%
  \def\AND{\end{tabular}\hfil\linebreak[4]\hfil
          \begin{tabular}[t]{c}\rule{\z@}{24pt}\ignorespaces}%
  {\centering
   \begin{tabular}[t]{c}\rule{\z@}{24pt}\@author\end{tabular}\par}
\else
  \lhead{Under review as a conference paper at ICLR 2026}
  \def\And{\end{tabular}\hfil\linebreak[0]\hfil
          \begin{tabular}[t]{c}\rule{\z@}{24pt}\ignorespaces}%
  \def\AND{\end{tabular}\hfil\linebreak[4]\hfil
          \begin{tabular}[t]{c}\rule{\z@}{24pt}\ignorespaces}%
  {\centering
   \begin{tabular}[t]{c}\rule{\z@}{24pt}Anonymous authors\\Paper under double-blind review\end{tabular}\par}
\fi
\vskip 0.3in minus 0.1in}}
\makeatother

\author{Xinjie Zhu\thanks{Xinjie Zhu and Zijing Zhao contributed equally to this work.} \\
{\small Lenovo Research, Beijing, China} \\
\texttt{\small zhuxj11@lenovo.com}
\And
Zijing Zhao\footnotemark[1] \\
{\small Lenovo Research, Beijing, China} \\
\texttt{\small zhaozj23@lenovo.com}
\AND
Hui Jin \\
{\small Lenovo Research, Beijing, China} \\
\texttt{\small jinhui8@lenovo.com}
\And
Qingxiao Guo \\
{\small Lenovo Research, Beijing, China} \\
\texttt{\small guoqx2@lenovo.com}
\And
Yilong Ma \\
{\small Lenovo Research, Beijing, China} \\
\texttt{\small mayl9@lenovo.com}
\And
Yunhao Wang \\
{\small Lenovo Research, Beijing, China} \\
\texttt{\small wangyh43@lenovo.com}
\And
Xiaobing Guo \\
{\small Lenovo Research, Beijing, China} \\
\texttt{\small guoxba@lenovo.com}
\And
Weifeng Zhang\thanks{Corresponding author: Weifeng Zhang (\texttt{weifengz@lenovo.com}).} \\
{\small Lenovo Research, Beijing, China} \\
\texttt{\small weifengz@lenovo.com}
}

\begin{document}

\maketitle

\begin{abstract}
Artificial Intelligence Generated Content (AIGC), particularly video generation with diffusion models, has been advanced rapidly. 
Invisible watermarking is a key technology for protecting AI-generated videos and tracing harmful content, and thus plays a crucial role in AI safety.
Beyond post-processing watermarks which inevitably degrade video quality, recent studies have proposed distortion-free in-generation watermarking for video diffusion models.
However, existing in-generation approaches are non-blind: they require maintaining all the message-key pairs and performing template-based matching during extraction, which incurs prohibitive computational costs at scale.
Moreover, when applied to modern video diffusion models with causal 3D Variational Autoencoders (VAEs), their robustness against temporal disturbance becomes extremely weak.
To overcome these challenges, we propose SIGMark, a Scalable In-Generation watermarking framework with blind extraction for video diffusion.
To achieve blind-extraction, we propose to generate watermarked initial noise using a Global set of Frame-wise PseudoRandom Coding keys (GF-PRC), reducing the cost of storing large-scale information while preserving noise distribution and diversity for distortion-free watermarking.
To enhance robustness, we further design a Segment Group-Ordering module (SGO) tailored to causal 3D VAEs, ensuring robust watermark inversion during extraction under temporal disturbance.
Comprehensive experiments on modern diffusion models show that SIGMark achieves very high bit-accuracy during extraction under both temporal and spatial disturbances with minimal overhead, demonstrating its scalability and robustness.
Our code is available at https://github.com/JeremyZhao1998/SIGMark-release.
\end{abstract}

\section{Introduction}
\label{sec:intro}

% I will write introduction here. 
% I will use url command like: \url{https://openreview.net/}.
% I will use inline file name like: \verb+iclr2026_conference.pdf+.
% I will use section reference like section~\ref{sec:intro}.
% I will use footnotes like: in this way\footnote{This way is how we use footnotes}.
% \input{figures/example}
% I will ref the figure like Figure~\ref{fig:example}.

In the field of Artificial Intelligence Generated Content (AIGC), diffusion models have rapidly advanced image and video generation \citep{croitoru2023diffusion,cao2024survey}. 
Latent diffusion models proposed by \cite{rombach2022high} generates images by denoising sampled noise in latent space. 
Extending from this, video diffusion models generate temporally coherent frame sequences by enforcing both spatial and temporal consistency \citep{ho2022video}. 
With the rapid proliferation of AI-generated videos, privacy and security concerns have become increasingly critical \citep{wang2024security}. 
On the one hand, as a widely used creative tool, AI-generated high-quality videos constitute valuable intellectual property (IP) and necessitate reliable copyright identification. 
On the other hand, the ease of producing harmful or misleading content calls for strict control mechanisms, requiring effective methods to trace their source of generation.

\begin{figure}[t]
    \begin{center}
        \includegraphics[width=1.0\linewidth]{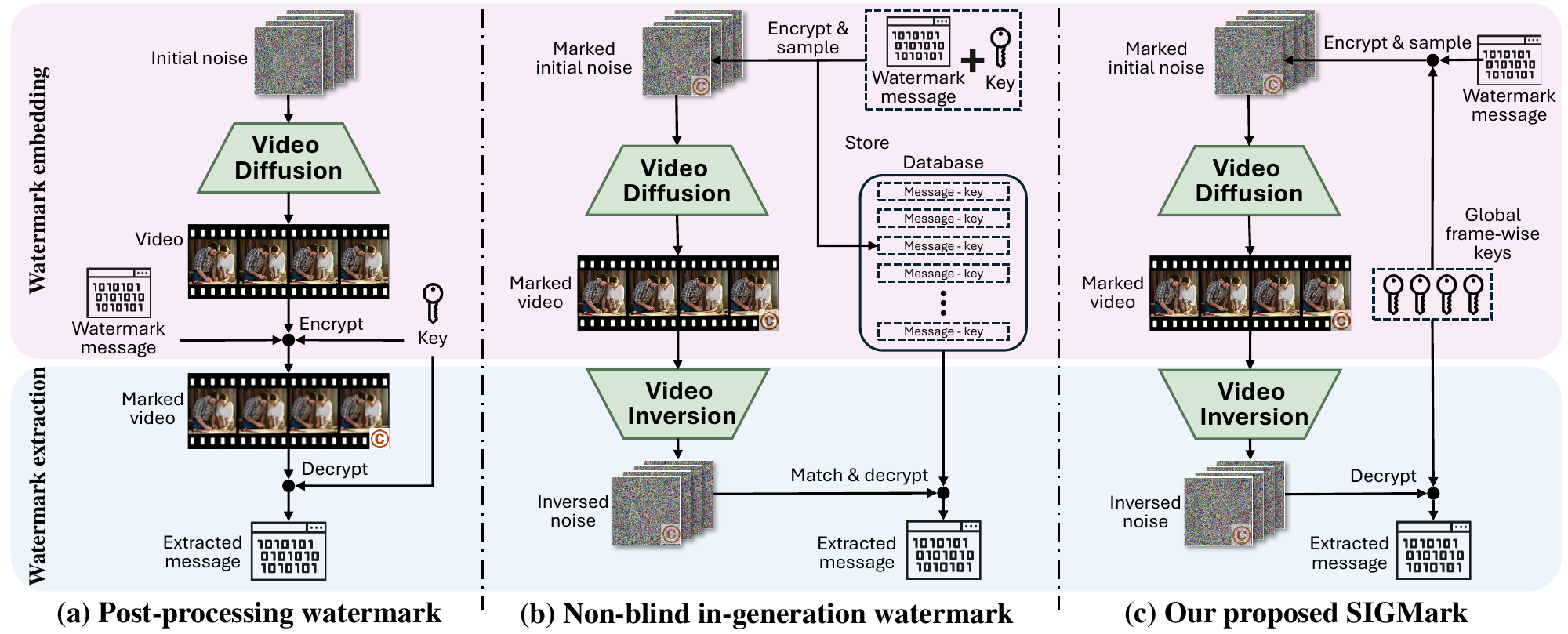}
    \end{center}
    \vspace{-3mm}
    \caption{
        (a) Post-processing watermarks: embedding watermarks in pixel-space which inevitably degrades video quality.
        (b) Existing in-generation methods: maintaining all the message-key pairs for matching, incurring high extraction costs and poor robustness.
        (c) Our proposed SIGMark: a blind watermarking framework with global frame-wise PRC keys which is inherently scalable.
    }
    \label{fig:teaser}
\end{figure}

To meet the demands of privacy and security, watermarking technology \citep{cox2007digital} has been widely applied in AIGC \citep{luo2025digital}. 
Invisible watermarking embeds information in a way imperceptible to the human eye, thereby preserving visual quality while remaining robust to various distortions and even malicious attacks \citep{wang2023data}. 
For AI-generated videos, a straightforward approach is to treat them like conventional videos and apply invisible watermarking to each frame after generation \citep{luo2023dvmark,zhang2024hideandtrack}, known as post-processing watermarking (see \Cref{fig:teaser}(a)).
However, such methods inevitably introduce redundant information, thereby degrading overall video quality. 
Recently, in-generation watermarking has been explored for both image \citep{yang2024gaussian,li2025gaussmarker} and video generation \citep{hu2025videoshield,hu2025videomark}. 
As illustrated in \Cref{fig:teaser}(b), these methods embed watermark messages during generation process, typically by sampling a watermarked initial noise in which the message is encoded using a secret key. 
During extraction, the watermarked video is inverted (commonly via Denoising Diffusion Implicit Models (DDIM) inversion) back into latent noise, and then decoded with the key to recover the watermark. 
These approaches have been theoretically proven to be distortion-free for diffusion models.

Although in-generation video watermarking offers the advantage of being distortion-free, it still faces two critical challenges:
(1) \textit{High extraction cost at scale.}
When extracting watermark messages from videos subject to temporal disturbances (e.g., frame removal during compression), current methods rely on template matching with the original latent noise. 
This requires maintaining all message-key pairs during watermark embedding, and computing matching functions across the entire database, as shown in \Cref{fig:teaser}(b). 
Such approaches are non-blind, with the extraction cost growing linearly with the scale of users or generation requests, severely limiting scalability.
(2) \textit{Poor temporal robustness.}
Modern video diffusion models \citep{yang2025cogvideox,kong2024hunyuanvideo} employ causal 3D Variational Autoencoders (VAE), which decode a group of adjacent frames from one temporal dimension of latent features. 
During extraction, video inversion requires the correct grouping of frames to reconstruct the latent feature. 
Temporal disturbance which disrupts the grouping will produce unrelated latent features, ultimately leading to very low extracted bit accuracy.

To address these issues and ensure usability at large-scale video generation platforms, we introduce SIGMark: scalable in-generation watermarking with blind extraction for video diffusion models.

To reduce the \textit{high extraction cost at scale}, we propose Global Frame-wise PseudoRandom Coding (GF-PRC) scheme for watermark embedding, thereby enabling a blind watermarking framework. 
Specifically, a global set of frame-wise PRC keys (pseudorandom error-correction code by \citet{christ2024pseudorandom}) is shared across all generation requests, with each key assigned to a group of temporal frames. 
Watermark messages are encoded into random latent noise with these keys, which preserves diversity and generative performance. 
During extraction, the global frame-wise PRC directly decodes the inverted noise without matching with the original messages. 
Throughout this process, the system only needs to maintain the global frame-wise PRC keys, reducing extraction complexity from linear in the number of generation requests to constant, and achieving strong scalability.

To enhance \textit{temporal robustness}, we introduce a Segment Group-Ordering (SGO) module tailored to causal 3D VAEs in modern video diffusion models. 
Specifically, for a video potentially affected by temporal disturbances, we first partition it into motion-consistent segments using Farnebäck optical flow; within each segment, a sliding-window grouping detector infers the original causal frame groups. 
This procedure recovers the correct grouping and, in turn, yields accurate inverted latents for watermark extraction under temporal disturbances.

We conduct comprehensive experiments on modern video diffusion models (HunyuanVideo by \citet{kong2024hunyuanvideo} and Wan-2.2 by \citet{wan2025wan}), covering both text-to-video (T2V) and image-to-video (I2V) pipelines. 
A subset of VBench-2.0\citep{zheng2025vbench20} under 18 evaluation dimensions is sampled to generate 400 videos for evaluation. 
Results show that our method achieves very high bit accuracy with high watermark capacity with minimal extraction cost. 
Our method maintains high accuracy under spatial and temporal disturbances, demonstrating strong robustness.
Our code is available at https://github.com/JeremyZhao1998/SIGMark-release.

In conclusion, the main contributions of this paper are:
\begin{itemize}
    \item We identify two critical issues in existing in-generation video watermarks: high extraction cost and poor temporal robustness, which hinder their scalability to large platforms.
    \item We propose SIGMark, a scalable in-generation watermarking framework with blind extraction for video diffusion, effectively addressing the limitations of scalability and robustness.
    \item We conduct extensive experiments for SIGMark on modern video diffusion models and comprehensive evaluation benchmarks, demonstrating its effectiveness and robustness.
\end{itemize}

\section{Related works}
\label{sec:related}

% When I cite a work using the author's names, I will use \verb+\citet{}+, like: \citet{Bengio+chapter2007} and \citet{Hinton06} introduce deep learning.
% When I only mention work, I will use \verb+\citep{}+, like: Deep learning is very useful \citep{Bengio+chapter2007,goodfellow2016deep}.

\subsection{Diffusion models}

In the field of Artificial Intelligence Generated Content (AIGC), diffusion models have rapidly advanced image and video generation.
Latent diffusion model (LDM)\citep{rombach2022high} synthesizes content by denoising sampled noise in latent space and decoding it through a Variational Autoencoder (VAE). 
Building on LDM, works such as SDXL~\citep{podell2024sdxl}, ControlNet~\citep{zhang2023adding} and DiT~\citep{peebles2023scalable} further improve image generation with stronger photorealism and higher resolution, fine-grained structural control, and greater scalability. 
Extending from images, video diffusion models tackle the task of generating temporally coherent frame sequences \citep{ho2022video}. 
Following the success of Sora~\citep{openai2024sora}, a wave of open-source video diffusion models including KLING~\citep{kuaishou2024kling}, HunyuanVideo~\citep{kong2024hunyuanvideo}, and Wan~\citep{wan2025wan} has emerged. 
They adopt causal 3D VAEs that compress videos along spatial and temporal dimensions to form compact latent sequences for diffusion, enabling longer, higher-quality videos with improved temporal consistency.
Our research focuses on watermarking for diffusion-generated videos and evaluates on HunyuanVideo and Wan-2.2. 

\subsection{Video watermarking}

Video invisible watermarking embeds imperceptible, durable signals in video to enable rights management and piracy deterrence \citep{asikuzzaman2017overview,aberna2024digital}. 
The straightforward image-based watermarking approaches operate frames individually \citep{hartung1998watermarking,hernandez2000dct}, while video-based works explicitly exploit temporal information via compressed domain \citep{biswas2005adaptive,noorkami2007framework} and motion vectors produced during compression \citep{mohaghegh2008h}. 
Recently, deep learning has been applied to watermarking for images \citep{zhu2018hidden} and videos \citep{ben2024deep}: 
\citet{zhang2019rivagan} introduced RivaGAN, training an encoder-decoder framework for robust watermarking. 
Subsequent work further advances robustness and capacity through curriculum learning \citep{ke2022robust}, low-order recursive Zernike-moment embedding \citep{he2023robust}, multiscale distribution modeling \citep{luo2023dvmark}, complex wavelet transforms \citep{yasen2025dtcwt,huang2025rbmark}, adversarially optimization under frequency domain \cite{huang2024robin}, and spatiotemporal attention \citep{li2024robust,yan2025spatiotemporal}.
However, embedding extra signals inevitably degrades visual quality.
We instead pursue in-generation watermarking for diffusion models, which is training-free and provably distortion-free.

\subsection{In-generation watermarking for diffusion models}

With the rapid progress of image and video diffusion models, watermarking for diffusion-generated content has likewise gained traction. 
Recent work integrates watermark embedding into the generative process to reduce performance degradation. 
\citet{fernandez2023stable} fine-tune the LDM decoder using a pre-trained watermark extractor, enabling reliable extraction from images produced by the fine-tuned model. 
\citet{yang2024gaussian} introduce Gaussian Shading, the first approach to sample watermarked initial noise for image generation, which is provably distortion-free. 
Subsequent studies further enhance robustness during the embedding and inversion phases \citep{li2025gaussmarker,fang2025cosda}. 
In-generation watermarking has also been extended from images to videos: \citet{liu2025implanting} propose a two-stage implanting scheme during the diffusion process. 
Other works encrypt the watermark message into the initial latent noise via Gaussian sampling \citep{hu2025videoshield} or dynamic tree-ring \citep{zeng2025dtr}, preserving video generation quality. 
\citet{hu2025videomark} adopt PRC for watermark encryption and decryption, maintaining generative diversity across messages.
Despite being distortion-free, advanced in-generation methods by \citet{hu2025videoshield,hu2025videomark} still face high extraction costs at scale and poor temporal robustness.
We are the first to identify and address these issues in modern video diffusion models, achieving strong scalability and temporal robustness.
\section{Method}
\label{sec:method}

\subsection{Problem formulation}

This paper focuses on in-generation watermarking for video diffusion models.
We first formalize the problem. 
Given a diffusion model $\mathcal{M}$, our goal is to embed a watermark message $m$ into generated video frames $\mathrm{VF}(m)=\mathrm{Embed}(\mathcal{M};m)$ without degrading the performance of the diffusion model. 
The watermarked video frames may undergo temporal disturbances (e.g., frame drops) or spatial disturbances (e.g., cropping), yet the tampered frames $\mathrm{VF}'$ should still permit reliable watermark extraction. 
We adopt the blind-watermark setting: during extraction, neither the original message $m$ nor the original generated video frames $\mathrm{VF}$ is available, and the message is recovered solely from the tampered video frames and the model, i.e., $\hat{m}=\mathrm{Extract}(\mathcal{M};\mathrm{VF}')$. 
We emphasize robustness: even under strong disturbances to $\mathrm{VF}'$, the recovered message $\hat{m}$ remains close to $m$.

During watermark embedding, we follow the in-generation scheme of \citet{yang2024gaussian,hu2025videoshield}. 
The watermark message $m$ is encrypted into the initial latent noise without altering its distribution, i.e., $z_0(m)\sim\mathcal{N}(0,\mathbf{I})$.
The model then denoises this noise with text prompts to generate videos, thereby preserving generative performance. 
However, existing methods require the original message $m$ for template matching during extraction, limiting scalability in real-world deployments. 
Moreover, when applied to modern diffusion models with causal 3D VAEs, they exhibit poor robustness under temporal disturbances.

\subsection{Framework overview}

\begin{figure}[t]
    \begin{center}
        \includegraphics[width=1.0\linewidth]{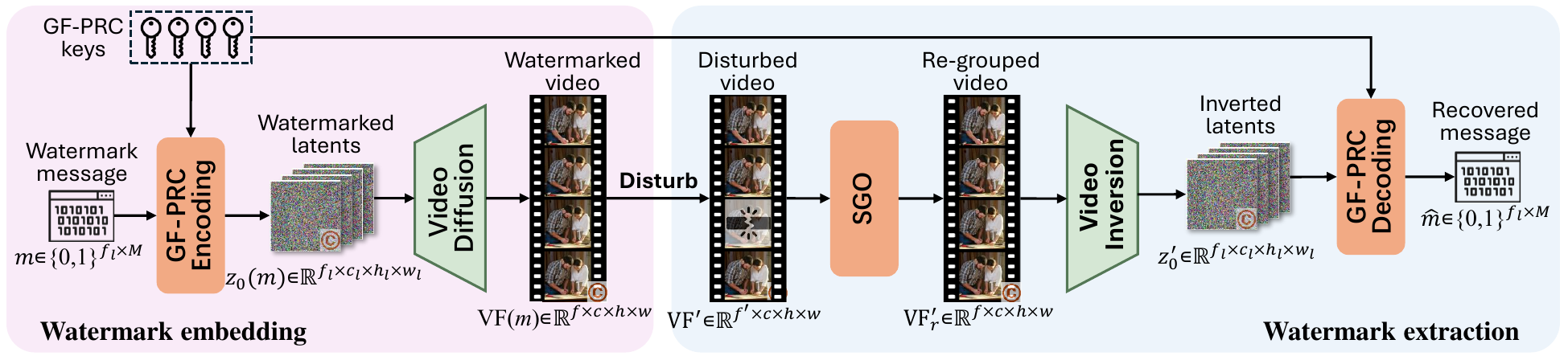}
    \end{center}
    \caption{
        Overview of our proposed SIGMark.
        Embedding: We encode the watermark message into the initial latent noise using a Global set of Frame-wise Pseudo-Random Coding (GF-PRC) keys. 
        The diffusion model then denoises this noise into video frames that carry the embedded messages.
        Extraction: A (possibly disturbed) video is first processed by our proposed Segment Group-Ordering (SGO) module to recover the correct causal frame grouping, then inverted to obtain the latent noise, from which the message is decoded using the GF-PRC keys.
        The system stores only the GF-PRC keys for both embedding and extraction, enabling blind watermarking.
    }
    \label{fig:method-overview}
\end{figure}

An overview of our scalable in-generation watermarking framework, SIGMark, is shown in \Cref{fig:method-overview}. 
We follow the in-generation watermarking scheme which embeds the watermark message into the initial latent noise to generate distortion-free watermarked video through diffusion (left part of \Cref{fig:method-overview}), and then process the video through inversion to obtain inverted latent noise for watermark extraction (right part of \Cref{fig:method-overview}).
To enable blind watermarking and reduce extraction cost at scale, we introduce a Global Frame-wise Pseudo-Random Coding (GF-PRC) scheme for message encryption and decryption. 
During embedding, the watermark message is encoded into the initial latent noise using a global set of frame-wise PRC keys, each key assigned to one temporal dimension of latent features. 
During extraction, given a possibly tampered video, a Segment Group-Ordering (SGO) module restores the correct causal frame grouping by Farnebäck optical flow segmentation and sliding-window grouping detector.
We then perform inversion to recover the watermarked latent and decode the message with the GF-PRC keys. 
GF-PRC enables blind extraction where only the global keys are stored in the system, while maintaining high-accuracy message recovery under disturbances.
We elaborate the details of our proposed modules in the following sections.

\subsection{Watermark embedding}

Following the in-generation watermarking scheme\citep{yang2024gaussian,hu2025videoshield}, we map the watermark message bits into the initial latent noise without affecting the Gaussian distribution of the noise for watermark embedding.
To achieve blind watermarking, we propose to utilize a Global set of Frame-wise Pseudo-Random coding (GF-PRC) scheme for watermark embedding.

\subsubsection{Video generation by modern diffusion models}

For a modern video diffusion model\citep{kong2024hunyuanvideo,wan2025wan} $\mathcal{M}$, which consists of a text prompt encoder $E_{\mathrm{text}}$, a denoising diffusion transformer $T$, and the encoder $E_{\mathrm{3D}}$ and decoder $D_{\mathrm{3D}}$ of the causal 3D Variational Autoencoders (VAE), the denoising steps happen on latent-space features $z\in \mathbb{R}^{f_l \times c_l \times h_l \times w_l}$, where $f_l, c_l, h_l, w_l$ denote the frame, channel, height, and width dimensions in latent space. 
The diffusion transformer $T$ denoises a randomly initialized latent noise $z_0\sim\mathcal{N}(o, \mathbf{I})$ guided by the text prompt:
$z_{\tau}=\mathrm{Denoise}\big(T;z_o;E_{\mathrm{text}}(\mathrm{prompt})\big)$,
where $\tau$ denotes the number of denoising steps. 
The denoised latent feature $z_\tau$ is then processed by the causal 3D VAE decoder to generate video frames $\mathrm{VF} \in \mathbb{R}^{f\times c\times h\times w}$, where $f, c, h, w$ denote the frame, channel, height, and width of the generated video. The whole process can be formulated as:
\begin{align}
    \mathrm{VF}&=\mathrm{Diffusion}(\mathcal{M};z_o;\mathrm{prompt})\\
    &=D_{\mathrm{3D}}\Big(\mathrm{Denoise}\big(T;z_o;E_{\mathrm{text}}(\mathrm{prompt})\big)\Big).
\end{align}
The causal 3D VAE introduces information compression along both spatial and temporal dimensions, where $f=f_l\times d_t$, $h = h_l \times d_s$, and $w = w_l \times d_s$, with $d_t, d_s$ denoting the temporal and spatial compression ratios, respectively. 
As a result, a group of $d_t$ frames is decoded from one temporal dimension of the latent features. 
Our proposed SIGMark embeds the watermark message into the latent noise via the GF-PRC scheme, yielding a sequence of watermark bits carried by a causal group of video frames, as detailed in the next paragraph.

\subsubsection{Global Frame-wise PseudoRandom Coding scheme}

\begin{figure}[t]
    \begin{center}
        \includegraphics[width=1.0\linewidth]{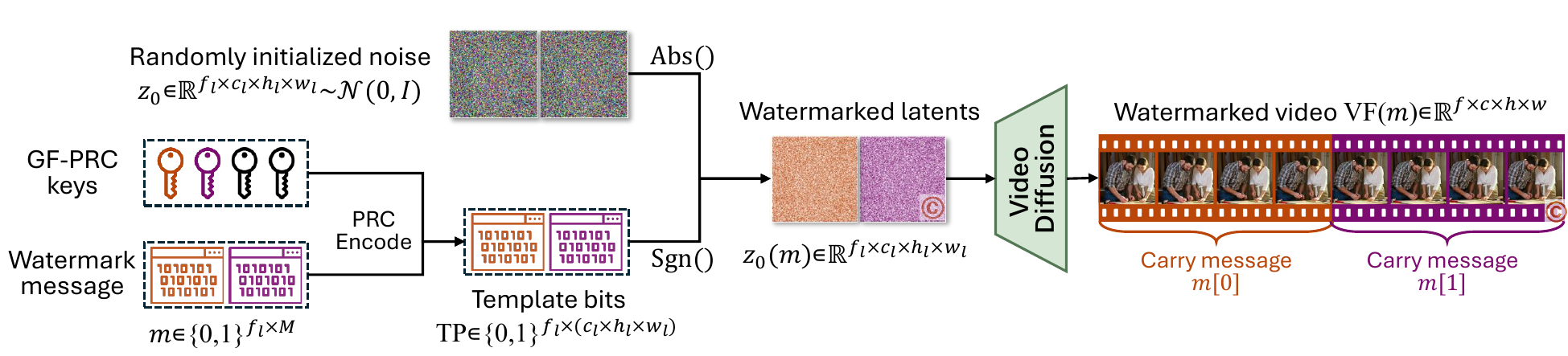}
    \end{center}
    \caption{
        Watermark embedding with GF-PRC scheme.
        We set $f_l=2, f=8$ with compression ratio $d_t=f/f_l=4$ as an example.
        \textcolor{orange}{Orange} and \textcolor{violet}{purple} denote message $m[0]$ and $m[1]$ respectively.
    }
    \label{fig:method-gf-prc}
\end{figure}

We propose embedding all watermark messages using a global set of frame-wise pseudorandom coding keys, as shown in \Cref{fig:method-gf-prc}. 
Specifically, for a watermark message $m\in \{0,1\}^{f_l\times M}$ where $f_l$ is the number of latent frames and $M$ is the bit length carried by each causal frame group, we encode $m$ into a random template bit sequence $\mathrm{TP}\in \{0,1\}^{f_l\times (c_l\times h_l \times w_l)}$ using the pseudorandom error-correction code (simplified as pseudorandom code, PRC) proposed by \citet{christ2024pseudorandom}:
\begin{align}
    \mathrm{TP}[i] = \mathrm{PRC.Encode}\big(m[i];K[i]\big), ~~i\in \{0,1,2,...,f_l-1\}
\end{align}
Here, $K$ denotes the pseudorandom error-correction coding keys; $K[i]$ is the key for frame dimension $i$ in latent space, with $m[i]\in \{0,1\}^{M}$ and $\mathrm{TP}[i]\in \{0,1\}^{(c_l\times h_l\times w_l)}$. 
We allocate one PRC key per causal frame group in latent space, enhancing robustness and enabling causal frame grouping and ordering information recovery (detailed in the next section).
Given the randomized template $\mathrm{TP}$, we map the watermark message into the initial latent noise by element-wise modulation:
\begin{align}
    z_0(m) = (\mathrm{TP} * 2 - 1) * |z_0|
\end{align}
With the random absolute value from Gaussian sampling and the randomized template bits as the modulation signal, the embedded initial noise remains Gaussian, $z_0(m)\sim\mathcal{N}(0,1)$, and thus does not degrade the diffusion model’s generative performance. 
Consequently, the generated video frames are watermarked with $m$ via:
\begin{align}
    \mathrm{VF}(m) = \mathrm{Diffusion}\big(\mathcal{M};z_0(m);\mathrm{prompt}\big)
\end{align}
Note that in our GF-PRC scheme, the frame-wise PRC keys are global: every generation request shares the same key set, and each latent frame dimension carries one watermark sequence encoded by its corresponding PRC key. 
The total number of GF-PRC keys can be set to the maximum frame capacity of the video generation system, enabling watermarking for videos of arbitrary length. 
PRC by \citet{christ2024pseudorandom} introduces a pseudo-random mapping that can encode even the same message into different random template bits, thereby preserving randomness in the initial latent noise under global keys, which traditional stream ciphers (e.g., ChaCha20\citep{bernstein2008chacha}) used in prior in-generation non-blind methods\citep{yang2024gaussian,hu2025videoshield} cannot provide with the fixed keying material.
The detailed explanation can be seen in Appendix A.

\subsection{Watermark extraction}

As shown in \Cref{fig:method-overview}, given a test video, we first apply the Segment Group-Ordering module to recover the grouping and ordering information of video frames, enhancing robustness against temporal disturbances, then perform diffusion-based inversion to obtain the inverted latent noise, and finally apply PRC decoding to recover the message.

\subsubsection{Segment Group-Ordering module}

\begin{figure}[t]
    \begin{center}
        \includegraphics[width=0.98\linewidth]{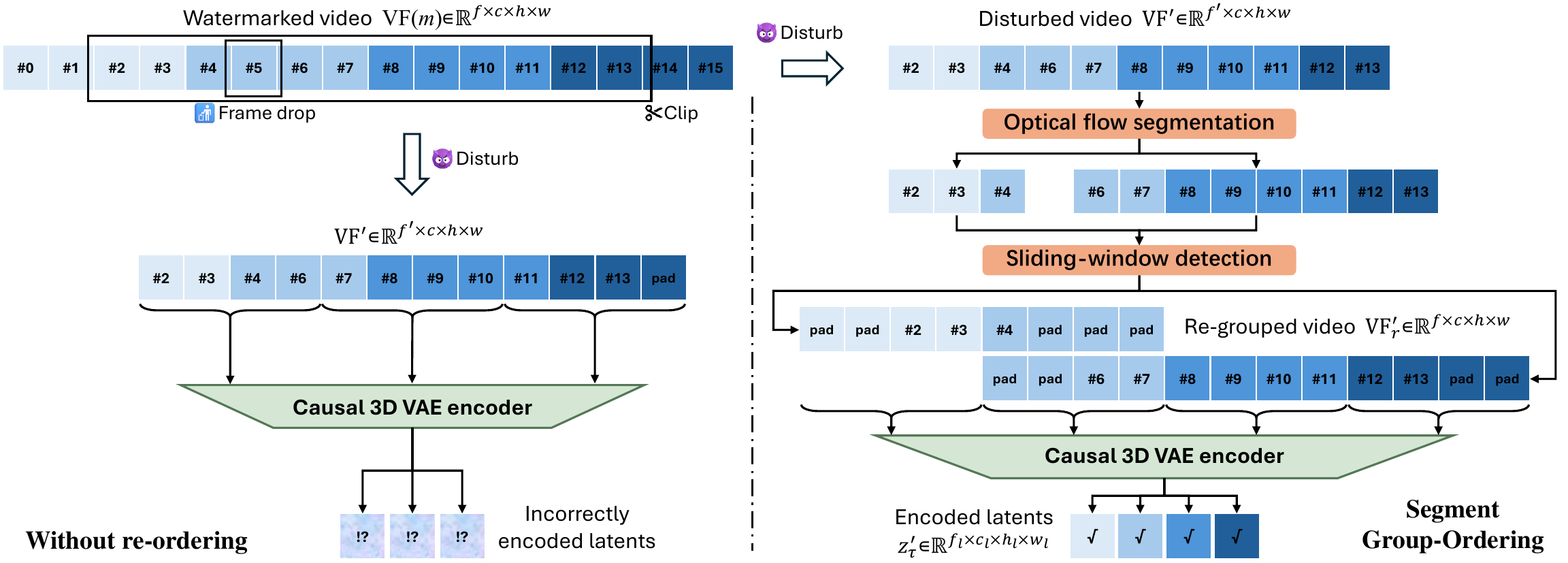}
    \end{center}
    \caption{
        Segment Group-Ordering (SGO) module.
        We set compression ratio $d_t=f/f_l=4$ as an example.
        When temporal disturbances (e.g., clipping or frame drops) occur, the causal grouping is disrupted; without re-ordering, this leads to incorrectly encoded latent features. 
        Our SGO module restores the correct grouping and ordering, yielding robust latent features for video inversion.
    }
    \label{fig:method-sgo}
\end{figure}

A watermarked video may encounter various disturbances, among which temporal disturbances are particularly critical for modern diffusion models with causal 3D VAEs. 
As shown on the left of \Cref{fig:method-sgo}, the original frames $\mathrm{VF}(m)$ are generated in causal groups of 4 (frames sharing the same color). 
After video clipping or frame drops, the disturbed video $\mathrm{VF'}$ loses both grouping and ordering information. 
If such frames are fed directly into the causal 3D VAE encoder, the mis-grouping (e.g., differently colored frames within a group) produces latent features that are inconsistent with those of $\mathrm{VF}(m)$, thereby preventing reliable recovery of the message bits via inversion.
To address the issue of weak temporal robustness, we propose a Segment Group-Ordering (SGO) module as shown in the right part of \Cref{fig:method-sgo} to recover the grouping and ordering information of the disturbed video for correct encoding.

\textbf{Optical flow segmentation:}
We propose a optical-flow method to partition a video into maximally contiguous subsequences with consistent temporal dynamics. 
For each adjacent pair $(\mathrm{VF}[t],\mathrm{VF}[t+1])$, we compute bidirectional Farnebäck flow and derive three indicators of temporal consistency: (i) median flow magnitude, (ii) forward-backward consistency, and (iii) motion-compensated residual.
These are normalized via median absolute deviation and combined with a weighted sum to form a discontinuity score. 
The score is smoothed with a Gaussian filter, and temporal cut points are detected using hysteresis thresholding. 
The procedure runs near real time and outputs contiguous frame segments.
Within each segment, we then perform sliding-window grouping detection. 
Further details of the optical-flow segmentation algorithm are provided in Appendix C.

\textbf{Sliding-window detection:}
Given a contiguous frame segment, we only need to identify the first frame of a causal group, and the subsequent frames can then be grouped correctly. 
To this end, we leverage the global frame-wise PRC keys.
Specifically, we pad $(d_t - 1)$ frames at the beginning of the segment and apply a sliding window over the padded sequence.
At sliding index $j$, we invert frames $[j:j+2*d_t]$ to obtain two latent frame dimensions $z_0'[j]$ and $z_0'[j+1]$.
Using the global PRC keys, the frame index of each latent can be determined by:
\begin{align}
    \hat{\mathrm{Idx}[j]} = \mathrm{argmax}\big(\mathrm{PRC.Detect}(z_0'[j];K[0,1,...,f_l])\big), \hat{\mathrm{Idx}[j]}\in \{0,1,...,f_l\}
\end{align}
We stop sliding when the detections are consecutive, i.e., $\hat{\mathrm{Idx}[j]} + 1 = \hat{\mathrm{Idx}[j+1]}$, indicating that the current grouping yields the correct segment start for inversion.
The results from all segments are then merged to produce the re-grouped frames $\mathrm{VF}_r'$, as shown in \Cref{fig:method-sgo}. 
Note that empty slots within each group are filled preferentially with available frames; any remaining slots are then padded.
If an entire group contains no frames (caused by frame drop), we assign padding based on the nearest available frame.
This procedure recovers correct ordering and grouping information and is robust to frame insertion, swapping, dropping, and clipping. 
Further details of the sliding-window detection algorithm are provided in Appendix C.

\subsubsection{Message bits extraction}

Following the in-generation watermarking paradigm\citep{yang2024gaussian,hu2025videoshield}, we perform inversion on the re-grouped frames to extract the watermark bits. 
The latent features $z_\tau'\in \mathbb{R}^{f_l\times c_l\times h_l\times w_l}$ are first obtained via the causal 3D VAE encoder, and then inverted through the diffusion transformer (the flow-matching Euler discrete inversion for HunyuanVideo\citep{kong2024hunyuanvideo} and Wan\citep{wan2025wan}):
\begin{align}
    z_\tau'&=E_{\mathrm{3D}}(\mathrm{VF}_r') \\
    z_0'&=\mathrm{Inversion}(\mathcal{M};z_\tau';\mathrm{prompt}_\emptyset)
\end{align}
where $\mathrm{prompt}_\emptyset$ denotes an empty prompt, since no original generation information is available. 
Given the inverted latent noise $z_0'\in \mathbb{R}^{f_l\times c_l\times h_l\times w_l}$, we decode the GF-PRC keys $K$ by applying PRC to the signs of $z_0'$:
\begin{align}
    \hat{m[i]} = \mathrm{PRC.Decode}\left(\frac{\mathrm{Sgn}\big(z_0'[i]\big) +1}{2};K[i]\right), ~~i\in \{0,1,2,...,f_l-1\}
\end{align}
where $\mathrm{Sgn}(\cdot)$ is the sign function. 
With the global PRC key set, no additional information needs to be stored during generation, and no template matching (as in \citet{hu2025videoshield,hu2025videomark}) is required during extraction.
Therefore, our proposed SIGMark enables blind watermarking with minor computation cost and strong scalability.
Details of message extraction cost are provided in Appendix B.

\section{Experiments}
\label{sec:exp}

In this section, we conduct experiments on modern video diffusion models. 
We detail the experimental setup in \Cref{sec:exp:settings}, report bit accuracy comparisons with existing methods in \Cref{sec:exp:results}, assess temporal and spatial robustness in \Cref{sec:exp:robustness}, conduct ablation studies in \Cref{sec:exp:ablation} and present evidence of scalability in \Cref{sec:exp:timecost}.

\begin{table}[t]
    \caption{Video watermarking results: message recovery bit accuracy and video quality score.\tablefootnote{We implement DCT and DT-CWT through the open-source code of \href{https://pypi.org/project/video-invisible-watermark/}{video-invisible-watermark} and \href{https://pypi.org/project/blind-video-watermark}{blind-video-watermark} respectively, and we implement VideoMark and VideoShield by adapting their officially released code and hyper-parameter to new diffusion models and prompts.}}
    \label{tab:main-results}
    \small
    \setlength{\tabcolsep}{2.5pt}
    \begin{center}
        \begin{tabular}{cc|cc|cc|cc|cc}
            \toprule
            \multicolumn{2}{c}{Diffusion model} & \multicolumn{4}{c}{HunyuanVideo T2V} & \multicolumn{4}{c}{HunyuanVideo I2V} \\
            \midrule
            \multicolumn{2}{c|}{Watermarking} & \multicolumn{2}{c|}{512 bits} & \multicolumn{2}{c|}{512$\times$16 bits} & \multicolumn{2}{c|}{512 bits} & \multicolumn{2}{c}{512$\times$16 bits} \\
            Method & Category & Bit acc$\uparrow$ & V-score$\uparrow$ & Bit acc$\uparrow$ & V-score$\uparrow$ & Bit acc$\uparrow$ & V-score$\uparrow$ & Bit acc$\uparrow$ & V-score$\uparrow$ \\
            \midrule
            No-mark & -- & -- & 0.490 & -- & 0.490 & -- & 0.463 & -- & 0.463 \\
            \midrule
            DCT & Post & 0.889 & 0.424 & 0.862 & 0.423 & 0.890 & 0.452 & 0.858 & 0.456 \\
            DT-CWT & Post & 0.619 & 0.416 & 0.650 & 0.436 & 0.627 & 0.458 & 0.611 & 0.463 \\
            VideoMark & None-blind & 0.873 & 0.507 & 0.758 & 0.502 & 0.846 & 0.483 & 0.707 & 0.482 \\
            VideoShield & None-blind & 1.000 & 0.497 & 0.991 & 0.506 & 1.000 & 0.482 & 0.999 & 0.482 \\
            \midrule
            SIGMark(Ours) & Blind & 0.958 & 0.506 & 0.885 & 0.499 & 0.981 & 0.472 & 0.905 & 0.488 \\
            \bottomrule
        \end{tabular}
    \end{center}
\end{table}

\subsection{Experimental settings}
\label{sec:exp:settings}

\subsubsection{Datasets and metric}

To comprehensively assess modern video diffusion models, we conduct experiments on a subset of prompts from VBench-2.0\citep{zheng2025vbench20}, which offers more evaluation dimensions and more complex prompts than VBench-1.0\citep{huang2023vbench} used by existing researches on earlier diffusion models\citep{hu2025videoshield,hu2025videomark}. 
Among the 18 dimensions in VBench-2.0, we select 3 prompts for the “diversity” dimension and generate 20 videos per prompt; for the remaining 17 dimensions, we select 5 prompts each and generate 4 videos per prompt, yielding a total of 400 videos. 
We report bit accuracy between the embedded message $m$ and the recovered message $\hat{m}$. 
Video quality is evaluated using the VBench-2.0 protocols to obtain an overall quality score. 
Detailed examples of the constructed prompt set are provided in Appendix D.

\subsubsection{Implementation details}

We evaluate on two open-source video diffusion models: HunyuanVideo\citep{kong2024hunyuanvideo} and Wan-2.2\citep{wan2025wan}, both employing causal 3D VAEs with temporal and spatial compression ratios $d_t=4$ and $d_s=8$. 
For each model, we consider both text-to-video (T2V) and image-to-video (I2V) tasks. 
For T2V, we generate videos at resolution $h=512$, $w=512$, with $65$ frames per video. 
The first frame is processed independently by the diffusion model, so no watermark is embedded in the first frame. 
The remaining $f=64$ frames form $f_l=f/d_t=16$ causal groups, for each group we maintain one frame-wise PRC key globally.
For I2V, we use the same text prompt set to generate the initial image prompt by FLUX\citep{blackforestlabs2025flux1} for evaluation.
The generated image prompt examples are shown in Appendix D.

\subsection{Watermark extraction results}
\label{sec:exp:results}

In this section, we compare our watermarking performance with existing approaches. 
As presented in \Cref{tab:main-results}, we report message bit accuracy (``Bit Acc'') and the VBench-2.0 score (``V-score'') for HunyuanVideo. 
Additional results for Wan-2.2 are provided in Appendix D.
We consider two configurations: (1) embedding an identical 512-bit watermark message in every causal frame group of a video, and (2) embedding a distinct 512-bit watermark message per causal frame group, corresponding to total capacities of 512 bits and 512×16=8192 bits per video, respectively. 

We compare our method against four baselines: DCT\citep{hartung1998dct}, DT-CWT\citep{coria2008dtcwt}, VideoShield\citep{hu2025videoshield}, and VideoMark\citep{hu2025videomark}. 
DCT and DT-CWT are widely used post-processing watermarking methods for general videos, whereas VideoShield and VideoMark are recent in-generation methods for video diffusion models but are non-blind. 
For post-processing methods, we first generate a set of standard videos with the diffusion model and then apply watermarking of different settings. 
As shown in \Cref{tab:main-results}, post-processing watermarking causes notable quality degradation relative to no-watermark videos, while in-generation methods leave visual quality essentially unaffected. 
Our proposed SIGMark can be proved to be performance-lossless.
Detailed proof can be found in Appendix A.
Our proposed SIGMark with blind extraction attains very high bit accuracy under both lower and higher capacity settings, surpassing VideoMark by large margins and remaining competitive with VideoShield which requires access to the original watermark information, thereby demonstrating the effectiveness of our approach.

\begin{table}[t]
    \caption{Watermark extraction bit accuracy under disturbances on HunyuanVideo I2V.}
    \label{tab:robustness}
    \small
    \begin{center}
        \begin{tabular}{c|cccc|cccc}
\toprule
\multirow{2}{*}{Method} & \multicolumn{4}{c}{Spatial disturbance} & \multicolumn{4}{|c}{Temporal disturbance} \\
 & w/o & G.noise & cmprs & blur & w/o & drop & insert & clip \\
\midrule
VideoMark 
& 0.85 
& 0.64$_{\downarrow 0.21}$ 
& 0.63$_{\downarrow 0.22}$ 
& 0.64$_{\downarrow 0.21}$ 
& 0.71 
& 0.52$_{\downarrow 0.19}$ 
& 0.51$_{\downarrow 0.20}$ 
& 0.51$_{\downarrow 0.20}$ 
\\
VideoShield
& 1.00
& 1.00$_{\downarrow 0.00}$
& 0.99$_{\downarrow 0.01}$
& 1.00$_{\downarrow 0.00}$
& 0.99
& 0.89$_{\downarrow 0.10}$
& 0.84$_{\downarrow 0.15}$
& 0.83$_{\downarrow 0.16}$
\\
SIGMark(Ours)
& 0.98
& 0.89$_{\downarrow 0.09}$
& 0.84$_{\downarrow 0.14}$
& 0.95$_{\downarrow 0.03}$
& 0.91
& 0.81$_{\downarrow 0.10}$
& 0.87$_{\downarrow 0.04}$
& 0.85$_{\downarrow 0.06}$
\\
\bottomrule
\end{tabular}
    \end{center}
\end{table}

\subsection{Robustness}
\label{sec:exp:robustness}

We assess the robustness of watermarking methods under both spatial and temporal perturbations in \Cref{tab:robustness}.
We apply spatial disturbance under 512 bits watermark and temporal disturbances under 512x16 bits. 
``w/o'' denotes without disturbance. ``G.noise'', ``cmprs'' and ``blur'' denotes Gaussian noise, mpeg compression and blurring, respectively. In temporal disturbances, we randomly drop, insert or clip out 30 frames. The performance degradation compared with no disturbances is marked as subscript.
For spatial disturbances, our approach incurs only marginal performance degradation, particularly when compared to VideoMark. 
Under the challenging temporal disturbances, VideoMark and VideoShield suffer substantial degradation on diffusion models with causal 3D VAEs, primarily due to incorrect grouping of causal frame groups. 
Our method mitigates this issue, achieving negligible performance loss and thereby improving temporal robustness.
It is worth noting that our method does not attain 100\% bit accuracy. 
We attribute this to the relationship of error-tolerance characteristics of PRC coding and the accuracy of diffusion inversion for different models. 
A detailed analysis of PRC coding robustness is provided in Appendix E.

\begin{table}[t]
    \caption{Ablation study on our proposed modules on HunyuanVideo I2V}
    \label{tab:ablation}
    \small
    \begin{center}
        \begin{tabular}{c|cc|cccc}
            \toprule
            \multirow{2}{*}{Method} & \multicolumn{2}{|c|}{Watermark embedding} & \multicolumn{4}{c}{Watermark extraction} \\
            & Single PRC & GF-PRC(Ours) & w/o SGO & w/o OF-seg & w/o SW-det & SGO(Ours) \\
            \midrule
            Bit acc & 0.707 & 0.905 & 0.534 & 0.762 & 0.823 & 0.869 \\
            \bottomrule
        \end{tabular}
        \vspace{-2mm}
    \end{center}
\end{table}

\subsection{Ablation studies}
\label{sec:exp:ablation}

We conduct ablation studies to evaluate the contribution of each proposed component in \Cref{tab:ablation}. 
``OF-seg'' and ``SW-det'' denotes optical flow segmentation and sliding window detection, respectively.
For the ablation study of watermark embedding modules, we evaluate a 512×16-bit watermark on generated videos without disturbances.
For the ablation study of watermark extraction modules, we evaluate a 512×16-bit watermark under temporal disturbance by inserting 30 random frames.
When SGO and OF-seg are removed, we simply truncate the video to the target length.
When SW-det is removed, we perform PRC detection by assuming that each segment starts at the beginning of a frame group.
For the GF-PRC scheme, removing it reduces our method to the same coding strategy as VideoMark, which exhibits limited bit accuracy under the blind setting without template matching. 
In contrast, GF-PRC not only enables blind extraction but also improves bit accuracy by introducing inter-frame redundant error tolerance. 
For the SGO module, both optical-flow-based segmentation and sliding-window detection are critical for recovering frame grouping and ordering; omitting either component leads to a noticeable accuracy drop.

\subsection{Evidence of Scalability}
\label{sec:exp:timecost}

\begin{figure}[t]
    \begin{center}
        \includegraphics[width=0.9\linewidth]{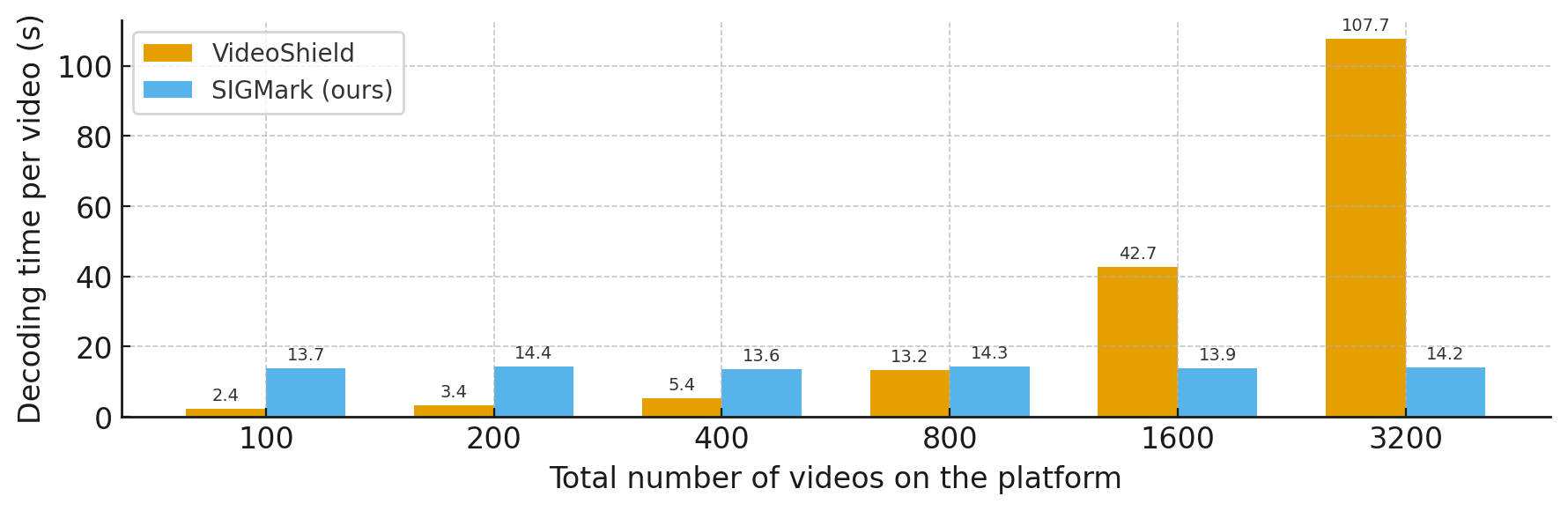}
    \end{center}
    \caption{
        The decoding time cost during watermark extraction. 
    }
    \label{fig:timecost}
\end{figure}

Baseline methods such as VideoShield and VideoMark are non-blind watermarking schemes: they require storing all watermark-related information (messages, encoding keys, etc.) during generation and matching against all stored information during extraction. Therefore, extraction cost grows with the total number of generated videos.
Our method, SIGMark, is blind: it maintains only a global set of frame-wise PRC keys and does not require any sample-specific metadata during extraction. As a result, it supports large-scale video generation platforms with constant extraction cost.
We analyze extraction time cost under scenarios where the total number of videos generated by the platform varies. 
Experiments are conducted under HunyuanVideo I2V with 512x16-bit watermark without disturbances. 
For fairness, we run inversion on GPU, and all remaining extraction steps including decryption and message matching on CPU. 
As shown in \Cref{fig:timecost}, the results demonstrate that the time cost of VideoShield scales linearly with the number of generated videos, which becomes impractical as the platform scales to millions of videos. In contrast, SIGMark remains constant, demonstrating strong scalability. 

\section{Conclusion}

Watermarking for video diffusion models is critical for ensuring safety and privacy control in AIGC. 
In this work, we introduce SIGMark, the first blind in-generation video watermarking method for modern diffusion models, offering strong scalability and practical applicability. 
To enable blind extraction without storing large-scale watermarking references, we propose a Global Frame-wise Pseudo-Random Code (GF-PRC) scheme, which encodes watermark messages into the initial latent noise without compromising video quality or diversity. 
To further enhance temporal robustness, we design a Segment Group-Ordering (SGO) module tailored for causal 3D VAEs, ensuring correct watermark inversion. 
Extensive experiments demonstrate that our approach achieves high bit accuracy with minimal overhead, validating its scalability and robustness.

% I will use unnumbered third level headings for the acknowledgments, and the Acknowledgments will only appear in camera-ready.

\bibliography{iclr2026_conference}
\bibliographystyle{iclr2026_conference}

\end{document}

\maketitle

% I will use unnumbered third level headings for the acknowledgments, and the Acknowledgments will only appear in camera-ready.

\appendix

In this material, we include the following contents as supplements of our paper:
(1) an explanation of our performance-lossless watermarking in \Cref{sec:performance-lossless};
(2) an analysis of the computational overhead of our method in \Cref{sec:computation-overhead}, demonstrating the scalability of SIGMark;
(3) detailed implementation of the proposed Segment Group-Ordering (SGO) module in \Cref{sec:detail-sgo};
(4) additional experimental details and results on video watermarking in \Cref{sec:additional-exp};
(5) an analysis of the performance bottlenecks of our method, outlining future research directions in \Cref{sec:limitations};
(6) a statement on the usage of Large Language Models (LLMs) in this paper in \Cref{sec:llm}.

\section{Impact of watermarking on video diffusion systems}
\label{sec:performance-lossless}

\subsection{Stream cipher unavailable for blind watermarking}

To embed a watermark message into the video diffusion model’s initial latent noise while remaining performance‑lossless, the message must be encoded via a cryptographic (pseudo) randomized encoder so that the induced perturbations are statistically indistinguishable from natural noise. 
Stream ciphers such as ChaCha20\citep{bernstein2008chacha} can produce pseudorandom code streams and offer good noise resilience.
However, their keystreams are deterministic given a fixed key (and nonce). Consequently, if a global key is reused and the message is fixed, the resulting codeword is also fixed, which harms generative diversity by mapping identical messages to identical latent sign patterns across runs. 
Preserving diversity would require a fresh per-generation secret (e.g., a new key/nonce) and thus auxiliary side information for extraction, rendering such schemes inherently none-blind and imposing substantial storage and coordination overhead at scale (tracking a large number of key-message pairs). 
Similar conclusion has been stated by \citet{thietke2025towards}.
In contrast, our approach targets blind watermarking with a single or a set of global keys by adopting a PRC (Pseudorandom error-correction Code proposed by \citet{christ2024pseudorandom}) that is explicitly randomized at encoding time: the same message admits many codewords drawn from a pseudorandom ensemble. 
This probabilistic encoding preserves sample diversity across generations without storing per-instance secrets, while still enabling blind extraction under global keys.

\subsection{Proof of our proposed method being performance-lossless}

Prior works often observe that inserting watermark embedding modules degrades model performance when evaluated by metrics such as Peak Signal-to-Noise Ratio (PSNR) and Fr\'echet Inception Distance (FID), which are more suitable for post-processing methods. 
To assess methods that integrate watermark embedding within the video generation process, we adopt a complexity-theoretic indistinguishability notion (inspired by cryptographic security definitions) to characterize the impact of watermarking on model performance. 
Concretely, consider a probabilistic game between a watermarked video $\mathrm{VF}(m)$ and a normally generated video $\mathrm{VF}$.

The watermarking method is \emph{performance-lossless} if, for any probabilistic polynomial-time (PPT) tester (distinguisher) $\mathsf{A}$, it holds that
\begin{equation}
\left|\Pr[\mathsf{A}(\mathrm{VF}(m))=1]-\Pr[\mathsf{A}(\mathrm{VF})=1]\right| < \operatorname{negl}(\rho)\,,
\tag{10}
\end{equation}
where $\rho$ is the security parameter (e.g., the length of the PRC keys $K$), and $\operatorname{negl}(\rho)$ denotes a function negligible in~$\rho$ (i.e., smaller than the inverse of any polynomial in~$\rho$).

We prove the claim by contradiction. Assume that $\mathrm{VF}(m)$ and $\mathrm{VF}$ are distinguishable by some PPT tester $\mathsf{A}$ with non-negligible advantage $\delta>0$:
\begin{equation}
\left|\Pr[\mathsf{A}(\mathrm{VF}(m))=1]-\Pr[\mathsf{A}(\mathrm{VF})=1]\right|=\delta\,.
\tag{11}
\end{equation}
Let the entire video generation pipeline, comprising the diffusion transformer $\mathrm{T}_{\mathrm{diff}}$ and the causal 3D VAE decoder $\mathrm{D}_{\mathrm{3D}}$, start from an initial latent noise $z^{(\cdot)}_0$. 
Substituting the generative process into~(11) yields
\begin{equation}
\left|\Pr\!\big[\mathsf{A}(\mathrm{D}_{\mathrm{3D}}(\mathrm{T}_{\mathrm{diff}}(z_0(m))))=1\big]
-\Pr\!\big[\mathsf{A}(\mathrm{D}_{\mathrm{3D}}(\mathrm{T}_{\mathrm{diff}}(z_0)))=1\big]\right|
=\delta\,,
\tag{12}
\end{equation}
where $z_0(m)$ denotes the initial latent noise whose sign pattern is determined by the message $m$ via our PRC encoding with keys $K$ (hence forming a pseudorandom sign sequence), and $z_0$ denotes the initial noise used for standard generation, drawn from a truly random distribution (e.g., $z_0 \sim \mathcal{N}(0,\mathbf{I})$).

Since $\mathrm{D}_{\mathrm{3D}}$ and $\mathrm{T}_{\mathrm{diff}}$ are deterministic PPT algorithms, they can be treated as subroutines available to the tester. 
Define a new tester $\mathsf{A}_{\mathrm{gen}} = \mathsf{A}\circ \mathrm{D}_{\mathrm{3D}}\circ \mathrm{T}_{\mathrm{diff}}$. 
Then~(12) reduces to distinguishing the initial latents:
\begin{equation}
\left|\Pr[\mathsf{A}_{\mathrm{gen}}(z_0(m))=1]-\Pr[\mathsf{A}_{\mathrm{gen}}(z_0)=1]\right|=\delta\,.
\tag{13}
\end{equation}
Equation~(13) asserts that a PPT algorithm can distinguish a latent tensor with a pseudorandom sign pattern ($z_0(m)$) from one with truly random signs ($z_0$) with non-negligible advantage.

However, our PRC mechanism is built upon cryptographic principles: the global keys $K$ are generated via a secure pseudorandom function (PRF). 
A fundamental property of a secure PRF is that its output is computationally indistinguishable from uniform randomness; consequently, the sign patterns produced by $\mathrm{PRC}.\mathrm{Encode}(K,m)$ are computationally indistinguishable from truly random signs. 
This contradicts~(13), which would enable distinguishing PRC-generated signs from random in PPT, violating the PRF security assumption.

Therefore, the assumption in~(11) is false. It follows that no PPT tester can distinguish watermarked videos from normally generated videos with non-negligible advantage, and thus our video watermarking method is performance-lossless under the above complexity-theoretic definition.

\section{Computation overhead}
\label{sec:computation-overhead}

\subsection{Computation complexity analysis}

Consider a large-scale video generation platform based on video diffusion models. We analyze the computational complexity of watermark embedding and extraction for our method and existing in-generation watermarking approaches~\citep{hu2025videoshield,hu2025videomark}.  
Let $f$ denote the maximum number of frames per video, $h \times w$ the spatial resolution, $t_{\mathrm{diff}}$ the time for denoising or inversion (correlated with $f \times h \times w$), $M$ the watermark message length, $K$ the key length, and $t_{\mathrm{encrypt}}$ the encryption/decryption time (correlated with $M$ and $f \times h \times w$). Let $N$ denote the number of videos or generation requests in the system.

For in-generation watermarking, the embedding process incurs no additional spatial cost beyond video generation, and the temporal cost is a single encryption step $t_{\mathrm{encrypt}}$, giving a total cost of: $t_{\mathrm{encrypt}} + t_{\mathrm{diff}}$.
However, non-blind watermarking requires storing key--message or key--template pairs for extraction, resulting in an additional spatial cost of:
$O(N \times (M + K))$.
For our proposed blind watermarking, only global frame-wise keys are maintained, with spatial cost:
$O(f \times K)$.
For extraction, non-blind watermarking requires matching against all stored watermark information. Under temporal disturbances, matching must be performed between watermark template bits for every frame, leading to a temporal complexity of: $O(N \times f \times f \times M)$, and a total complexity of:
$O(N \times f^2 \times M) + t_{\mathrm{diff}} + t_{\mathrm{decrypt}}$.
In contrast, our proposed SIGMark compares inverted latents with all frame-wise keys, yielding a total cost of: $t_{\mathrm{diff}} + f \times t_{\mathrm{decrypt}}$.

In typical experiments where $N$, $M$, and $f$ are of the same order, the dominant cost is $t_{\mathrm{diff}}$. However, in large-scale systems where $N \gg f$, the parameters $f,h,w,M,K$ can be treated as constants (e.g., $f=64$, $h=w=512$, $M=512$, while $N \sim 10^{8}$). In this setting, the $O(N)$ spatial and temporal cost of non-blind watermarking becomes prohibitive, whereas our proposed SIGMark maintains constant complexity, demonstrating strong scalability.

\subsection{Experimented computation time and memory usage on hardware}

All experiments are conducted on NVIDIA A800 GPUs with \texttt{bfloat16} precision. On HunyuanVideo, generating a single video with $f=65$, $h=512$, and $w=512$ takes approximately 4 minutes, with watermark embedding time being negligible. 
Processing the dataset of 400 videos requires about 1600 A800 GPU hours.  
For watermark extraction, PRC detection and decoding take roughly 30 seconds per sample on CPU, which can potentially be optimized through parallelization. Notably, this computation time remains constant regardless of the scale of video generation.

\section{Detailed implementation of SGO}
\label{sec:detail-sgo}

\subsection{Detailed implementation of Optical Flow segmentation}

We aim to partition a sequence of frames $\{I_t\}_{t=0}^{n-1}$ into contiguous segments that are likely temporally consistent, even under frame deletions, insertions, or swaps. 
For pre-processing, each frame is converted to grayscale and downscaled so that its short side does not exceed a fixed size (default 288), reducing computational cost without introducing upsampling artifacts.
For Optical Flow Estimation, for each adjacent pair $(I_t, I_{t+1})$, we compute forward and backward Farnebäck optical flow fields, denoted as $\mathbf{F}^{t\to t+1}$ and $\mathbf{F}^{t+1\to t}$.
We then calculate boundary features.
For each boundary $t|t+1$, we compute four robust signals:
\begin{align}
M_t &= \operatorname{median}_{\mathbf{x}}\;\bigl\lVert \mathbf{F}^{t\to t+1}(\mathbf{x}) \bigr\rVert_2, \tag{14}\\
C_t &= \operatorname{median}_{\mathbf{x}}\;\bigl\lVert 
\mathbf{F}^{t\to t+1}(\mathbf{x}) + \mathbf{F}^{t+1\to t}(\mathbf{x} + \mathbf{F}^{t\to t+1}(\mathbf{x})) \bigr\rVert_2, \tag{15}\\
R_t &= \frac{1}{255|\Omega|}\sum_{\mathbf{x}\in\Omega}\bigl| \hat{I}_{t\to t+1}(\mathbf{x}) - I_{t+1}(\mathbf{x}) \bigr|, \tag{16}\\
\Delta M_t &= 
\begin{cases}
0, & t=0,\\
|M_t - M_{t-1}|, & t\ge 1.
\end{cases} \tag{17}
\end{align}
Here, $M_t$ is the median flow magnitude, $C_t$ measures forward–backward consistency, $R_t$ is the motion-compensated residual after warping $I_t$ to $I_{t+1}$ using backward flow, and $\Delta M_t$ captures speed changes.
Each signal is standardized using a robust Z-score based on the median absolute deviation (MAD):
\begin{equation}
z(x) = 0.6745 \cdot \frac{x - \operatorname{median}(x)}{\operatorname{MAD}(x)+10^{-9}}. \tag{18}
\end{equation}
The final discontinuity score for boundary $t$ is a weighted combination:
\begin{equation}
\mathrm{score}_t = 0.35|z_M| + 0.30\max(z_C,0) + 0.25\max(z_R,0) + 0.10\max(z_{\Delta M},0). \tag{19}
\end{equation}
A light Gaussian smoothing is applied to reduce jitter.
We finally apply hysteresis thresholding with high and low thresholds $(\texttt{score\_hi}, \texttt{score\_lo})$ to detect cut points, reducing false positives from noise. Segments are then assembled as:
\[
[0, i_1], [i_1+1, i_2], \dots, [i_k+1, n-1].
\]
\textbf{Complexity.} The dominant cost is optical flow estimation for $n-1$ frame pairs, while memory overhead is minimal. The method is CPU-friendly and robust to noise, making it suitable for large-scale video analysis.

\subsection{Details of sliding-window detection}

The sliding-window detection module aims to determine both the grouping and the relative position of a given video segment within the entire generated video. Leveraging the detection capability of the GF-PRC scheme, we achieve this goal efficiently.
Given a video segment $\mathrm{VF}[t:t+l]$, we first pad the segment with $(c_t-1)$ frames at the beginning, where $c_t$ denotes the temporal compression ratio of the causal 3D VAE (e.g., $c_t=4$). We then apply a sliding window of size $2\times c_t$ (e.g., $8$) over the padded sequence. For each window index $j$, we perform inversion to obtain the local inverted initial noise:
\begin{equation}
z_\tau'[j:j+1] = E_{\mathrm{3D}}\bigl(\mathrm{VF}_r'[j:j+2\times c_t]\bigr), \tag{20}
\end{equation}
\begin{equation}
z_0'[j:j+1] = \mathrm{Inversion}\bigl(\mathcal{M}; z_\tau'[j:j+1]; \mathrm{prompt}_\emptyset\bigr). \tag{21}
\end{equation}
Next, we use the GF-PRC keys to detect whether $z_0'[j:j+1]$ matches any global frame index using Eq.~(6) from the main paper. This approach is effective because PRC detection accuracy is significantly higher than PRC decoding, as discussed in \Cref{sec:limitations}. We also observe that repeating the same frame $c_t$ times yields high inversion and detection accuracy, enabling the padding strategy to function correctly within the sliding window.
The sliding process terminates once the detected indices form a continuous sequence, as described in the main paper. The computational complexity of this procedure is at most $O(f \times c_t)$ and at least $O(f)$, which remains constant with respect to the overall generation scale.
\begin{table}[t]
    \caption{Our constructed prompt set -- Sample of 5 dimensions with representative prompts}
    \label{tab:example}
    \small
    \centering
    \setlength{\tabcolsep}{4pt} % 可按需要调小/调大列间距
    \renewcommand{\arraystretch}{1.15} % 行高，避免太挤

    % C / P / I 三列：都水平居中 + 垂直居中
    \begin{tabular}{
        >{\centering\arraybackslash}m{0.16\textwidth}
        >{\centering\arraybackslash}m{0.62\textwidth}
        >{\centering\arraybackslash}m{0.18\textwidth}
    }
        \toprule
        \textbf{Dimension} & \textbf{Prompt Text} & \textbf{Image} \\
        \midrule

        \textbf{Complex Plot} &
        The race commenced with explosive energy as the first runner from Team A surged ahead, capturing the attention of all spectators. His swift pace set the tone, but during the baton exchange, a slight fumble nearly cost them. Despite the near mishap, he managed to hand off the baton to the second runner, who, with fierce determination, pursued Team B relentlessly, overtaking them on the curve. However, the earlier error meant their lead was slim. As the third runner took over, nerves were palpable, yet he maintained their position, though Team C's strategic pacing brought them dangerously close. In the final handoff, the pressure was immense. The last runner from Team A accelerated on the penultimate turn, unleashing a powerful sprint that widened the gap. With unwavering focus, he crossed the finish line triumphantly, securing victory for Team A amidst roaring cheers. &
        \includegraphics[width=0.95\linewidth]{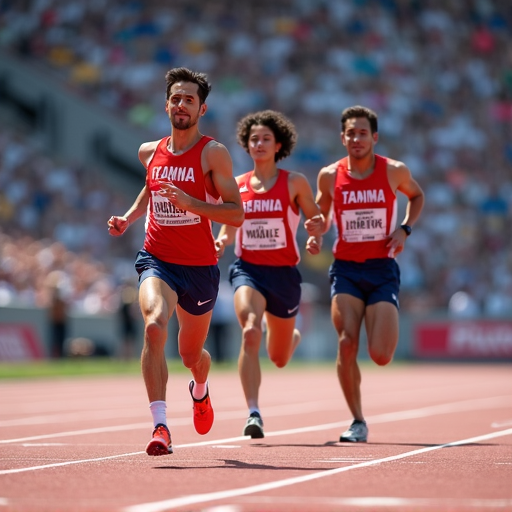} \\

        \midrule
        
        \textbf{Composition} &
        A magnificent lion, its golden mane flowing like a regal crown, possesses the expansive, powerful wings of an eagle, each feather shimmering under the sun's radiant glow. It soars effortlessly through a vast, azure sky, the clouds parting gracefully in its wake. The lion's eyes, sharp and focused, scan the horizon with a kingly gaze, while its muscular body glides with the grace of a seasoned aviator. Below, the earth stretches out in a patchwork of greens and browns, rivers glistening like silver ribbons. The scene captures a breathtaking blend of strength and elegance, as the majestic creature commands the heavens with unparalleled ease. &
        \includegraphics[width=0.95\linewidth]{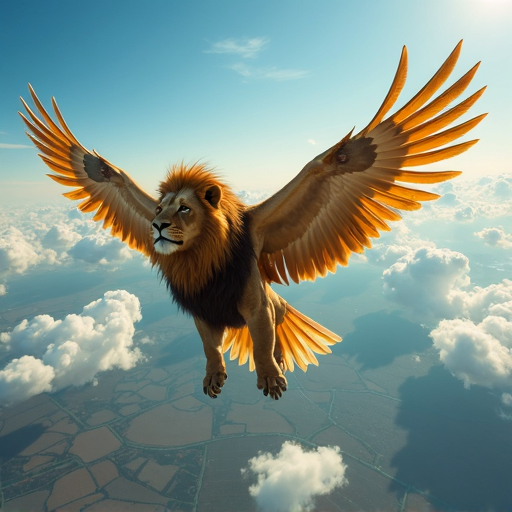} \\

        \midrule

        \textbf{Dynamic Attribute} &
        In a serene forest, the camera captures a mesmerizing transformation as vibrant red leaves slowly transition to lush green. The scene begins with a close-up of a single leaf, its deep crimson hue glowing under the gentle sunlight. As the camera pans out, the surrounding foliage reveals a tapestry of red, orange, and yellow, creating a warm, autumnal atmosphere. Gradually, the colors shift, with hints of green emerging, symbolizing the renewal of life. The sunlight filters through the canopy, casting dappled shadows on the forest floor, while a gentle breeze rustles the leaves, enhancing the sense of change and rebirth. &
        \includegraphics[width=0.95\linewidth]{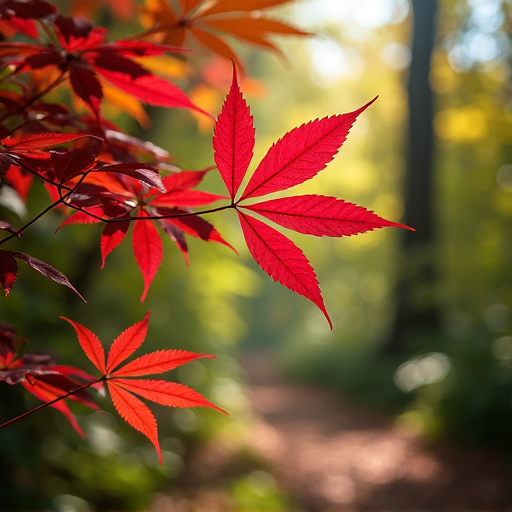} \\

        \midrule

        \textbf{Dynamic Spatial Relationship} &
        A playful dog, with its fur slightly ruffled by the breeze, stands on the left side of a wooden dining table, its ears perked up and eyes full of curiosity. The table is surrounded by cozy chairs and a vase with fresh flowers on top. As the dog gets excited, it begins to sprint energetically towards the front of the table, its paws making soft thudding sounds against the wooden floor, eager to reach the other side where a chew toy lies waiting. &
        \includegraphics[width=0.95\linewidth]{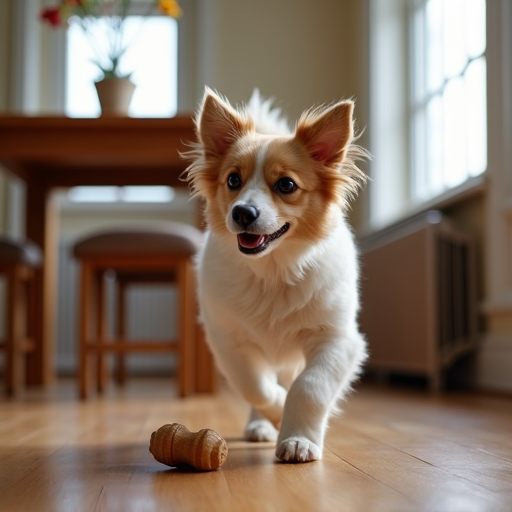} \\

        \midrule

        \textbf{Instance Preservation} &
        A vibrant orange dog, with a sleek coat and playful demeanor, sprints joyfully across a sunlit meadow, its ears flapping in the breeze. The dog's eyes sparkle with excitement as it bounds over the lush green grass, leaving a trail of paw prints behind. In the background, wildflowers sway gently, adding bursts of color to the scene. The sun casts a warm glow, highlighting the dog's energetic movements and the natural beauty surrounding it. As the dog runs, its tail wags enthusiastically, embodying pure joy and freedom in the open, expansive landscape. &
        \includegraphics[width=0.95\linewidth]{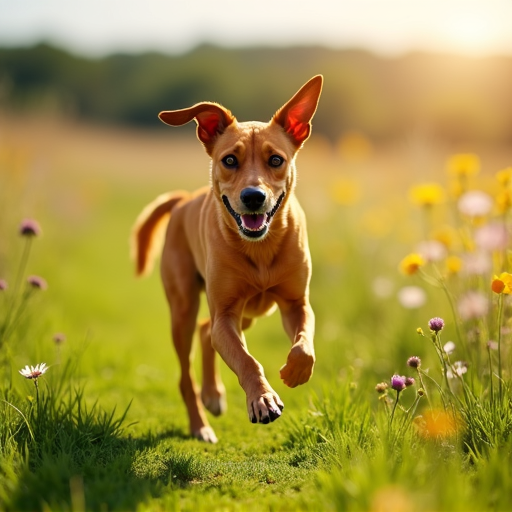} \\

        \bottomrule
    \end{tabular}
\end{table}

\begin{table}[t]
    \caption{Results of different models across different dimensions}
    \label{tab:wan_result}
    \small
    \begin{center}
        \begin{tabular}{c|cccc}
        \toprule
        Dimension & Hunyuan-I2V & Hunyuan-T2V & Wan-I2V & Wan-T2V \\
        \midrule
        complex plot                 & 1.0000 & 0.9292 & 1.0000 & 1.0000 \\
        composition                  & 1.0000 & 0.9313 & 1.0000 & 1.0000 \\
        dynamic attribute            & 1.0000 & 0.8878 & 1.0000 & 0.8974 \\
        dynamic spatial relationship & 0.9773 & 0.9991 & 1.0000 & 0.9554 \\
        instance preservation        & 0.9117 & 0.9584 & 1.0000 & 0.9150 \\
        motion rationality           & 0.9684 & 1.0000 & 1.0000 & 0.9379 \\
        multi-view consistency       & 0.9617 & 0.9707 & 0.9500 & 0.8820 \\
        camera motion                & 0.9560 & 0.9373 & 1.0000 & 1.0000 \\
        complex landscape            & 0.9913 & 1.0000 & 1.0000 & 1.0000 \\
        diversity                    & 1.0000 & 0.9794 & 1.0000 & 0.9695 \\
        human anatomy                & 0.9594 & 0.9787 & 1.0000 & 1.0000 \\
        human clothes                & 0.9879 & 0.9927 & 1.0000 & 0.9744 \\
        human identity               & 0.9969 & 0.9304 & 1.0000 & 0.9403 \\
        human interaction            & 1.0000 & 1.0000 & 1.0000 & 0.8864 \\
        material                     & 1.0000 & 0.9738 & 1.0000 & 0.8966 \\
        mechanics                    & 0.9684 & 0.9377 & 1.0000 & 1.0000 \\
        motion order understanding   & 0.9819 & 0.9617 & 1.0000 & 1.0000 \\
        thermotics                   & 0.9972 & 0.8815 & 0.9500 & 0.9360 \\
        \midrule
        Mean acc            & 0.9810 & 0.9583 & 0.9944 & 0.9550 \\
        \bottomrule
        \end{tabular}
    \end{center}
    
\end{table}

\section{Additional experimental results}
\label{sec:additional-exp}

\subsection{Our constructed prompt set}

To comprehensively assess both video quality and the efficiency of the watermarking framework, we adopt VBench-2.0~\citep{zheng2025vbench20}, a benchmark specifically designed for modern video diffusion models with stronger ability. VBench-2.0 evaluates 18 dimensions, including: \texttt{Camera Motion}, \texttt{Complex Landscape}, \texttt{Complex Plot}, \texttt{Composition}, \texttt{Diversity}, \texttt{Dynamic Attribute}, \texttt{Dynamic Spatial Relationship}, \texttt{Human Anatomy}, \texttt{Human Clothes}, \texttt{Human Identity}, \texttt{Human Interaction}, \texttt{Instance Preservation}, \texttt{Material}, \texttt{Mechanics}, \texttt{Motion Order Understanding}, \texttt{Motion Rationality}, \texttt{Multi-view Consistency}, and \texttt{Thermotics}.

We use the augmented text prompts curated for HunyuanVideo and other advanced video diffusion models. Compared to VBench-1.0 used by VideoShield~\citep{hu2025videoshield}, our selected prompt set is more comprehensive, featuring longer and more complex descriptions that better evaluate generative capability. In total, we select 88 text prompts and generate 400 videos for each diffusion model under evaluation. For image-to-video tasks, we employ FLUX-1.0~\citep{blackforestlabs2025flux1} to generate image prompts. Examples of text prompts and images used in our experiments are shown in \Cref{tab:example}. The full dataset will be released publicly.

\subsection{Results on Wan-2.2}

To further evaluate our approach, we performed additional experiments on the Image-to-Video (I2V) and Text-to-Video (T2V) variants of the Wan-2.2 model. The results demonstrate the significant effectiveness of our method, especially showcasing superior accuracy on the I2V model. We have meticulously recorded the performance across 18 distinct prompt dimensions, with the detailed outcomes presented in \Cref{tab:wan_result}.

\section{Performance bottleneck and future works}
\label{sec:limitations}

\begin{figure}[t]
    \begin{center}
        \includegraphics[width=1.0\linewidth]{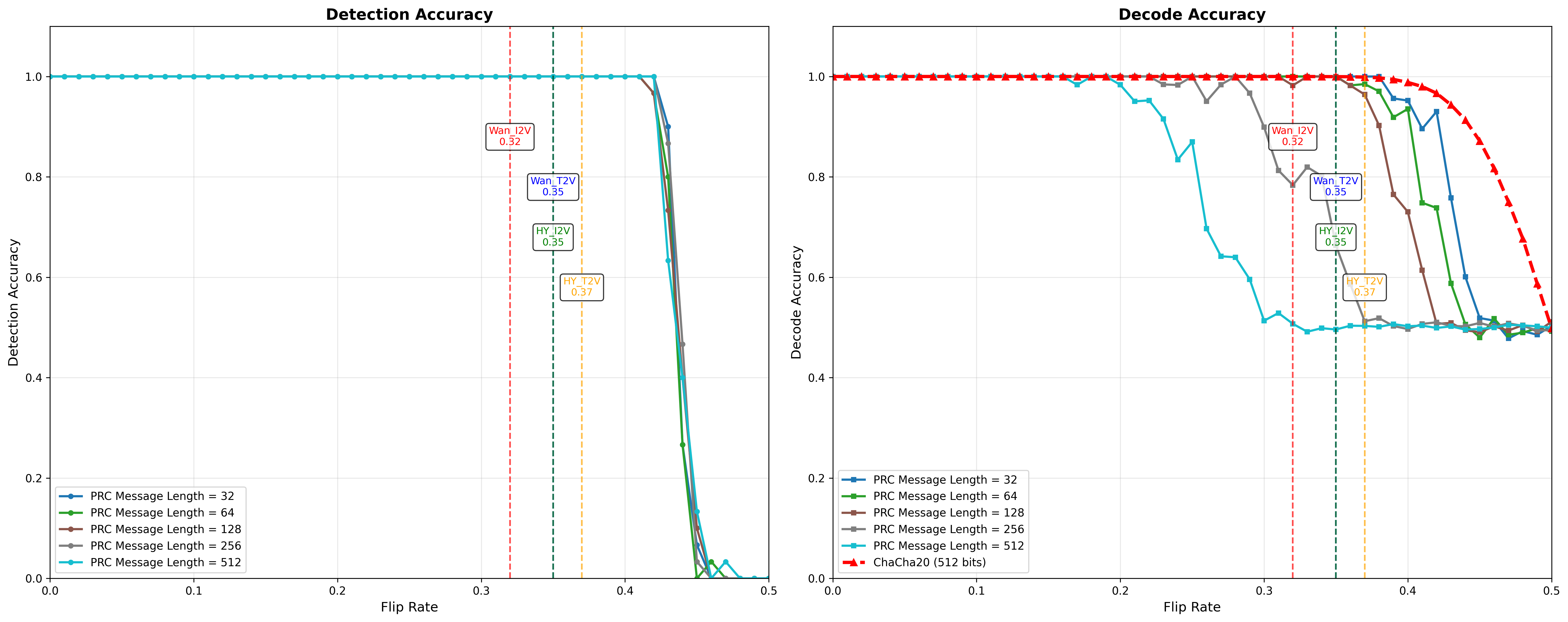}
    \end{center}
    \caption{
        PRC error-tolerance curve and the inversion accuracy of different video diffusion models.
    }
    \label{fig:prc-error}
\end{figure}

Watermarking methods such as VideoShield~\citep{hu2025videoshield} report nearly $100\%$ bit accuracy, but they are non-blind and incur high extraction costs at scale. In contrast, our method achieves high, though not perfect, extraction accuracy. We attribute this gap to two factors: (1) the imperfect inversion of video diffusion models and (2) the limited error-tolerance of PRC coding.

\Cref{fig:prc-error} illustrates the relationship between inversion error and the error-tolerance characteristics of PRC. While PRC introduces diversity by encoding messages into pseudorandom sequences, this design compromises its robustness to bit flips. In \Cref{fig:prc-error}, we plot PRC detection and decoding accuracy (Y-axis) under varying bit-flip rates (X-axis). When the error rate exceeds a certain threshold, PRC decoding accuracy drops sharply toward $0.5$ (random guessing). Moreover, as the encoded message length increases, this threshold becomes lower.

Inversion accuracy for current video diffusion models is far from perfect: HunyuanVideo exhibits an error rate of approximately $0.35$, while Wan-2.2 performs slightly better at $0.32$. When inversion error surpasses PRC’s tolerance, decoding becomes effectively random. In contrast, ChaCha20 combined with majority voting (red curve in \Cref{fig:prc-error}) demonstrates strong error resilience, which explains why non-blind methods like VideoShield achieve near-perfect extraction accuracy.

Fortunately, while PRC.decode() has limited fault tolerance capability, PRC.detect() exhibits strong robustness and has the potential to deliver reliable watermarking under extreme disturbances. We conduct experiments to test the false-positive rate of our proposed SIGMark, feeding the T2V generated videos into I2V model and PRC keys. The results show that 100\% of the false-positive samples are distinguished as not generated by I2V model, demonstrating that our method provide reliable watermarking.

Looking forward, we identify two promising directions to improve PRC-based blind watermarking: (1) enhancing inversion accuracy through improved video inversion techniques, and (2) introducing an additional error-correcting layer on top of PRC to mitigate inversion errors.

\section{Usage of Large Language Models}
\label{sec:llm}

This paper utilize Large Language Models (GPT-5) to aid the grammar mistakes and polish writing and formula, and it also helps in dealing with formatting issues during writing in \LaTeX.
We also use LLM-assisted coding systems in designing and implementing experiments.

\bibliography{iclr2026_conference}
\bibliographystyle{iclr2026_conference}